\documentclass[11pt,a4paper]{article}
\pdfoutput=1
\usepackage[hyperref]{emnlp2020}
\usepackage{times}
\usepackage{latexsym}
\usepackage{algorithmicx}
\usepackage{algpseudocode}
\usepackage{algorithm}
\usepackage{subcaption}
\usepackage[normalem]{ulem}

\usepackage{array}
\usepackage{svg}
\usepackage{soul}
\sethlcolor{pink}
\usepackage{fixltx2e}
\MakeRobust{\Call}
\usepackage{microtype}
\usepackage{multirow}
\usepackage{booktabs}
\usepackage{pgfplots}
\usepackage{amsmath}
\usepackage{enumitem}

\usepackage{todonotes}

\usepackage{multicol}
\pgfplotsset{every tick label/.append style={font=\small}}
\aclfinalcopy 

\usepackage[nameinlink]{cleveref}
\crefformat{subsection}{\S#2#1#3}
\Crefname{equation}{Eq.}{Eqs.}
\Crefname{figure}{Fig.}{Figs.}
\crefformat{subsubsection}{\S#2#1#3}

\usepackage{amsmath,amsfonts,bm}

\newcommand{\mor}{\textsc{Morpheus}}

\newcommand{\mytilde}{{\raise.17ex\hbox{$\scriptstyle\mathtt{\sim}$}}}
\newcommand{\rowgroup}[1]{\hspace{-1em}#1}

\def\eqref#1{equation~\ref{#1}}

\def\1{\bm{1}}

\DeclareMathAlphabet{\mathsfit}{\encodingdefault}{\sfdefault}{m}{sl}
\SetMathAlphabet{\mathsfit}{bold}{\encodingdefault}{\sfdefault}{bx}{n}

\title{Mind Your Inflections! Improving NLP for \\ Non-Standard Englishes with Base-Inflection Encoding}

\author{Samson Tan$^{\S\natural}$, Shafiq Joty$^{\S\ddagger}$, Lav R. Varshney$^{\mho\S}$, Min-Yen Kan$^\natural$\\
  $^\S$Salesforce AI Research\quad
  $^\natural$National University of Singapore \\
  $^\ddagger$Nanyang Technological University \quad
  $^\mho$University of Illinois at Urbana-Champaign\\
  $^\S$\texttt{\{samson.tan,sjoty\}@salesforce.com} \\ 
  $^\natural$\texttt{kanmy@comp.nus.edu.sg}\\
  $^\mho$\texttt{varshney@illinois.edu}
}

\date{}

\begin{document}
\maketitle
\begin{abstract}
Inflectional variation is a common feature of World Englishes such as Colloquial Singapore English and African American Vernacular English. Although comprehension by human readers is usually unimpaired by non-standard inflections, current NLP systems are not yet robust. We propose Base-Inflection Encoding (BITE), a method to tokenize English text by reducing inflected words to their base forms before reinjecting the grammatical information as special symbols. Fine-tuning pretrained NLP models for downstream tasks using our encoding defends against inflectional adversaries while maintaining performance on clean data. Models using BITE generalize better to dialects with non-standard inflections without explicit training and translation models converge faster when trained with BITE. Finally, we show that our encoding improves the vocabulary efficiency of popular data-driven subword tokenizers. Since there has been no prior work on quantitatively evaluating vocabulary efficiency, we propose metrics to do so.\footnote{Code will be available at \href{https://github.com/salesforce/bite}{github.com/salesforce/bite}.}
\end{abstract}

\section{Introduction}
Large-scale neural models have proven successful at a wide range of natural language processing (NLP) tasks but are susceptible to amplifying discrimination against minority linguistic communities \citep{hovy-spruit-2016-social,morpheus20} due to selection bias in the training data and model overamplification \citep{shah2019predictive}.

Most datasets implicitly assume a distribution of error-free Standard English speakers, but this does not accurately reflect the majority of the global English speaking population who are either second language (L2) or non-standard dialect speakers \citep{crystal2003english,ethno2019}. These World Englishes differ at lexical, morphological, and syntactic levels \citep{KachruKN2009}; sensitivity to these variations predisposes English NLP systems to discriminate against speakers of World Englishes by either misunderstanding or misinterpreting them \citep{translation-arrest2017,tatman-2017-gender}. Left unchecked, these biases could inadvertently propagate to future models via metrics built around pretrained models, such as BERTScore \citep{bert-score}.

In particular, \citet{morpheus20} show that current question answering and machine translation systems are overly sensitive to non-standard inflections---a common feature of dialects such as Colloquial Singapore English (CSE) and African American Vernacular English (AAVE).\footnote{Examples in \Cref{app:dialect}.} Since people naturally correct for or ignore non-standard inflection use \cite{FosterW2016}, we should expect NLP systems to be equally robust.

\begin{figure}[t]
    \centering
    \includegraphics[width=0.45\textwidth]{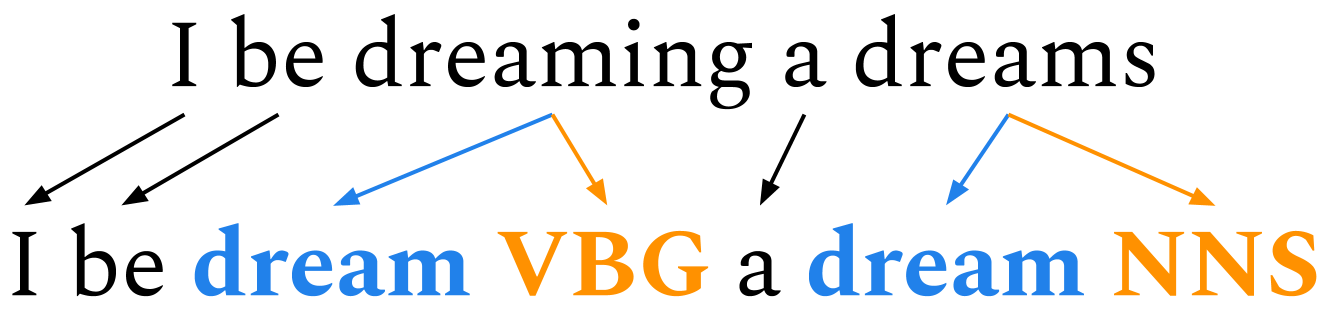}
    \caption{Base-Inflection Encoding reduces inflected words to their base forms, then reinjects the grammatical information into the sentence as inflection symbols.}
    \label{fig:base_inflection}
\end{figure}

Existing work on adversarial robustness for NLP primarily focuses on adversarial training methods \citep{belinkov2018synthetic,SinghGR18,morpheus20} or classifying and correcting adversarial examples \citep{zhou-etal-2019-learning}. However, this effectively increases the size of the training dataset by including adversarial examples or training a new model to identify and correct perturbations, thereby significantly increasing the overall computational cost of creating robust models.

These approaches also only operate on either raw text or the model, ignoring tokenization---an operation that transforms raw text into a form that the neural network can learn from. We introduce a new representation for word tokens that separates base from inflection. This improves both model robustness and vocabulary efficiency by explicitly inducing linguistic structure in the input to the NLP system \citep{erdmann-etal-2019-little,henderson2020unstoppable}.

Many extant NLP systems use a combination of a whitespace and punctuation tokenizer followed by a data-driven subword tokenizer such as byte pair encoding (BPE; \citet{sennrich2015neural}). However, a purely data-driven approach may fail to find the optimal encoding, both in terms of vocabulary efficiency and cross-dialectal generalization. This could make the neural model more vulnerable to inflectional perturbations. Hence, we:

\begin{itemize}[leftmargin=*,itemsep=0.5pt]
    \item Propose Base-InflecTion Encoding (BITE), which uses morphological information to help the data-driven tokenizer use its vocabulary efficiently and generate robust symbol\footnote{Following \citet{sennrich2015neural}, we use \emph{symbol} instead of \emph{token} to avoid confusion with the unencoded \emph{word token}.} sequences. In contrast to morphological segmentors such as Linguistica \citep{goldsmith2000linguistica} and Morfessor \citep{morfessor}, we reduce inflected forms to their base forms before reinjecting the inflection information into the encoded sequence as special symbols. This approach gracefully handles the canonicalization of words with nonconcatenative morphology while generally allowing the original sentence to be reconstructed.
    \item Demonstrate BITE's effectiveness at making neural NLP systems robust to non-standard inflection use while preserving performance on Standard English examples. Crucially, simply fine-tuning the pretrained model for the downstream task after adding BITE is sufficient. Unlike adversarial training, BITE does not enlarge the dataset and is 
    more computationally efficient.
    \item Show that BITE helps BERT \citep{bert19} generalize to dialects unseen during training and also helps Transformer-big \citep{ott2018-scaling} converge faster for the WMT'14 En-De task.
    \item Propose metrics like symbol complexity to operationalize and evaluate the vocabulary efficiency of an encoding scheme. Our metrics are generic and can be used to evaluate any tokenizer.
\end{itemize}

\section{Related Work}
\paragraph{Subword tokenization.}
Before neural models can learn, raw text must first be encoded into symbols with the help of a 
fixed-size vocabulary. Early models represented each word as a single symbol in the vocabulary \cite{neurallm,Collobert2011} and uncommon words were represented by an \texttt{unknown} symbol. However, such a representation is unable to adequately deal with words absent in the training vocabulary. Therefore, subword representations like WordPiece \cite{schuster2012wordpiece} and BPE \citep{sennrich2015neural} were proposed to encode out-of-vocabulary (OOV) words by segmenting them into subwords and encoding each subword as a separate symbol. This way, less information is lost in the encoding process since OOV words are approximated as a combination of subwords in the vocabulary. \citet{wang2019bbpe} reduce vocabulary sizes by operating on bytes instead of characters (as in standard BPE).

To make subword regularization more tractable, \citet{kudo-2018-subword} proposed an alternative method of building a subword vocabulary by reducing an initially oversized vocabulary down to the required size with the aid of a unigram language model, as opposed to incrementally building a vocabulary as in WordPiece and BPE variants. However, machine translation systems operating on subwords still have trouble translating rare words from highly-inflected categories \citep{koehn-knowles-2017-six}.

\citet{sadat-habash-2006-combination}, \citet{koehn-hoang-2007-factored}, and \citet{kann-schutze-2016-single} propose to improve machine translation and morphological reinflection by encoding morphological features separately while \citet{sylak-glassman-etal-2015-language} propose a schema for inflectional features. \citet{avraham-goldberg-2017-interplay} explore the effect of learning word embeddings from base forms and morphological tags for Hebrew, while \citet{chaudhary-etal-2018-adapting} show that representing words as base forms, phonemes, and morphological tags improve cross-lingual transfer for low-resource languages.

\paragraph{Adversarial robustness in NLP.}
To harden NLP systems against adversarial examples, existing work largely uses adversarial training \citep{goodfellow2015,jia-liang-2017-adversarial,ebrahimi-etal-2018-hotflip,belinkov2018synthetic,SinghGR18,iyyer-etal-2018-adversarial,cheng-etal-2019-robust}. However, this generally involves \emph{retraining} the model with the adversarial data, which is computationally expensive and time-consuming. \citet{morpheus20} showed that simply fine-tuning a trained model for a single epoch on appropriately generated adversarial training data is sufficient to harden the model against inflectional adversaries. 
Instead of adversarial training, \citet{moe} train word embeddings to be robust to misspellings, while \citet{zhou-etal-2019-disp} propose using a BERT-based model to detect adversaries and recover clean examples. \citet{jia-etal-2019-certified} and \citet{huang-etal-2019-achieving} use Interval Bound Propagation to train provably robust pre-Transformer models, while \citet{robusttrans20} propose an efficient algorithm for training certifiably robust Transformer architectures.

\paragraph{Summary.} Popular subword tokenizers operate on surface forms in a purely data-driven manner. Existing adversarial robustness methods for large-scale Transformers are computationally expensive, while provably robust methods have only been shown to work for pre-Transformer architectures and small-scale Transformers. 

Our work uses linguistic information (inflectional morphology) in conjunction with data-driven subword encoding schemes to make large-scale NLP models robust to non-standard inflections and generalize better to L2 and World Englishes, while preserving performance for Standard English. We also show that our method helps existing subword tokenizers use their vocabulary more efficiently.

\section{Linguistically-Grounded Tokenization}
Data-driven subword tokenizers like BPE improve a model's ability to approximate the semantics of unknown words by splitting them into subwords. 

Although the fully data-driven nature of such methods make them language-agnostic, this forces them to rely only on the statistics of the surface forms when transforming words into subwords since they do not exploit any language-specific morphological regularities. To illustrate, the past tense of \textit{go}, \textit{take}, and \textit{keep} have the inflected forms \textit{went}, \textit{took}, and \textit{kept}, respectively, which have little to no overlap with their base forms\footnote{Base (no quotes) is synonymous with lemma in this paper.} and each other even though they share the same tense. These six surface forms would likely have no subwords in common in the vocabulary.  Consequently, the neural model would have the burden of learning both the relation between base forms and inflected forms and the relation between inflections for the same tense. Additionally, since vocabularies are fixed before model training, such an encoding does not optimally use a limited vocabulary. 

Even when inflections do not orthographically alter the base form and there is a significant overlap between the base and inflected forms, e.g., the \textit{-ed} and \textit{-d} suffixes, the suffix may be encoded as a separate subword and  base forms / suffixes may not be consistently represented.  To illustrate, encoding \textit{danced} as $[\textit{dance}, \textit{d}]$ and \textit{dancing} as $[\textit{danc}, \textit{ing}]$ results in two different ``base forms'' for the same word, \textit{dance}. This again burdens the model with learning the two ``base forms'' mean the same thing and makes inefficient use of a limited vocabulary.

When encoded in conjunction with another inflected form like \textit{entered}, which should be encoded as $[\textit{enter}, \textit{ed}]$, this encoding scheme also produces two different subwords for the same type of inflection \textit{-ed} vs \textit{-d}. As in the first example, the burden of learning that the two suffixes correspond to the same tense is transferred to the learning model. 

A possible solution is to instead encode \textit{danced} as $[\textit{danc}, \textit{ed}]$ and \textit{dancing} as $[\textit{danc}, \textit{ing}]$, but there is no guarantee that a data-driven encoding scheme will learn this pattern without some language-specific linguistic supervision. In addition, this unnecessarily splits up the base form into two subwords \textit{danc} and \textit{e}; the latter contains no extra semantic or grammatical information yet increases the encoded sequence length. Although individually minor, encoding many base words in this manner increases the computational cost for any encoder or decoder network.

Finally, although it is theoretically possible to force a data-driven tokenizer to segment inflected forms into morphologically logical subwords by limiting the vocabulary size, many inflected forms are represented as individual symbols at common vocabulary sizes (30--40k).
We found that the BERT$_{base}$ WordPiece tokenizer and BPE\footnote{Trained on Wikipedia+BookCorpus (1M) with a vocabulary size of 30k symbols.} encoded each of the above examples as single symbols.

\begin{algorithm}[t]
\small
\begin{algorithmic}
\Require Input sentence $S=[w_1,\dots,w_N]$
\Ensure Encoded sequence $S'$
\State $S' \gets [\emptyset]$
\ForAll {$i = 1, \ldots, |N|$}
    \If{$\Call{POS}{w_i}\in \{$NOUN$,$ VERB$,$ ADJ$\}$}
        \State base $\gets $ \Call{GetLemma}{$w_i,$ \Call{POS}{$w_i$}}
        \State inflection $\gets$ \Call{GetInflection}{$w_i$}
        \State $S' \gets S' + [\text{base}, \text{inflection}]$
    \Else
        \State $S' \gets S' + [w_i]$
    \EndIf
\EndFor
\State \Return $S'$
\caption{Base-InflecTion Encoding (BITE)}
\label{alg:bite}
\end{algorithmic}
\end{algorithm}

\subsection{Base-Inflection Encoding}
To address these issues, we propose the Base-InflecTion Encoding framework (or BITE), which encodes the base form and inflection of content words separately. Similar to how existing subword encoding schemes improve the model's ability to approximate the semantics of out-of-vocabulary words with in-vocabulary subwords, BITE helps the model better handle out-of-distribution inflection usage by keeping a content word's base form consistent even when its inflected form drastically changes. This distributional deviation could manifest as adversarial examples, such as those generated by \mor\ \cite{morpheus20}, or sentences produced by L2 or World Englishes speakers. By keeping the base forms consistent, BITE provides adversarial robustness to the model.

\begin{table*}[t!]
    \small
    \centering
    \begin{tabular}{l c c c c c c c c}
    \toprule
    & \multicolumn{2}{c}{\textbf{SQuAD 2 Ans. (F$_1$)}} & \multicolumn{2}{c}{\textbf{SQuAD 2 All (F$_1$)}} & \multicolumn{2}{c}{\textbf{MNLI (Acc.)}} & \multicolumn{2}{c}{\textbf{MNLI-MM (Acc.)}} \\
    \textbf{Encoding} & Clean & \mor\ & Clean & \mor\ & Clean & \mor\ & Clean & \mor\ \\
    \midrule
    WordPiece (WP) & \textbf{74.58} & 61.37 & \textbf{72.75} & 59.32 & \textbf{83.44} & 58.70 & \textbf{83.59} & 59.75 \\
    BITE + WP & 74.50 & \textbf{71.33} & 72.71 & \textbf{69.23} & 83.01 & \textbf{76.11} & 83.50 & \textbf{76.64} \\
    \midrule
    WP + Adv. FT. & \textbf{79.07} & 72.21 & \textbf{74.45} & 68.23 & \textbf{83.86} & \textbf{83.87} & \textbf{83.86} & 75.77 \\
    BITE + WP (+1 epoch) & 75.46 & \textbf{72.56} & 73.69 & \textbf{70.66} & 82.21 & 81.05 & 83.36 & \textbf{81.04} \\
    \bottomrule
    \end{tabular}
    \caption{BERT$_{\text{base}}$ results on the clean and adversarial MultiNLI and SQuAD 2.0 examples. We compare BITE+WordPiece to both WordPiece alone and with one epoch of adversarial fine-tuning. For fair comparison with adversarial fine-tuning, we trained the BITE+WordPiece model for an extra epoch (bottom) on clean data.}
    \label{tab:adv_robustness}
    \end{table*}{}

\paragraph{BITE (\Cref{fig:base_inflection}).} 
Given an input sentence $S=[w_1,\dots,w_N]$ where $w_i$ is the $i^{th}$ word, BITE generates a sequence of symbols $S'=[w'_1,\dots,w'_N]$ such that $w_i' = [ \textsc{Base}(w_i)\text,  \textsc{Inflect}(w_i)]$ where $\textsc{Base}(w_i)$ is the base form of the word and $\textsc{Inflect}(w_i)$ is the inflection (grammatical category) of the word (\Cref{alg:bite}). If $w_i$ is not inflected, $\textsc{Inflect}(w_i)$ is $\textsc{Null}$ and excluded from the sequence of symbols to reduce the neural network's computational cost. In our implementation, we use Penn Treebank tags to represent inflections. 

By lemmatizing each inflected word to obtain the base form instead of segmenting it like in most data-driven encoding schemes, BITE ensures this base form is consistent for all inflected forms of a word, unlike a subword produced by segmentation, which can only contain characters present in the original word. For example, $\textsc{Base}(took)$, $\textsc{Base}(taking)$, and $\textsc{Base}(taken)$ all correspond to the same base form, \textit{take}, even though it is orthographically significantly different from \textit{took}.

Similarly, encoding all inflections of the same grammatical category (e.g., verb-past-tense) in a canonical form should help the model to learn each inflection's grammatical role more quickly. This is because the model does not need to first learn that the same grammatical category can manifest in orthographically different forms. 

Crucially, the original sentence can usually be reconstructed from the base forms and grammatical information preserved by the inflection symbols, except in cases of overabundance \citep{thornton2019overabundance}.

\paragraph{Implementation details.} We use the BertPreTokenizer from the \texttt{tokenizers}\footnote{\href{https://github.com/huggingface/tokenizers}{github.com/huggingface/tokenizers}} library for whitespace and punctuation splitting. We use the NLTK \citep{nltk09} implementation of the averaged perceptron tagger \citep{collins2002discriminative} with greedy decoding to generate POS tags, which serve to improve lemmatization accuracy and as inflection symbols. For lemmatization and reinflection, we use \texttt{lemminflect}\footnote{\href{https://github.com/bjascob/LemmInflect}{github.com/bjascob/LemmInflect}}, which uses a dictionary look-up together with rules for lemmatizing and inflecting words. A benefit of this approach is that the neural network can now generate orthographically appropriate inflected forms by generating the base form and the corresponding inflection symbol.

\subsection{Compatibility with Data-Driven Methods}
\label{sec:data-driven-compat}
Although BITE has the numerous advantages outlined above, it suffers from the same weakness as regular word-level tokenization schemes when used alone: a limited ability to handle out-of-vocabulary words. Hence, we designed BITE to be a general framework that seamlessly incorporates existing data-driven schemes to take advantage of their proven ability to handle OOV words.

To achieve this, a whitespace/punctuation-based pretokenizer is first used to transform the input into a sequence of words and punctuation characters, as is common in  machine translation. Next, BITE is applied and the resulting sequence is converted into a sequence of integers by a data-driven encoding scheme (\Cref{fig:bite_pipe} in \Cref{app:impl_details}). In our experiments, we use BITE in this manner and refer to the combined tokenizer as ``BITE+$D$", where $D$ refers to the data-driven encoding scheme.

\section{Model-Based Experiments}
We first demonstrate the effectiveness of BITE using the pretrained cased BERT$_{\text{base}}$ \citep{bert19} before training a full Transformer \citep{vaswani2017attention} from scratch. We do not replace WordPiece and BPE but instead incorporate them into the BITE framework as described in \Cref{sec:data-driven-compat}. The advantages and disadvantages to this approach will be discussed in the next section. We do not do any hyperparameter tuning but use the original models' in all experiments (detailed in \Cref{app:impl_details}).

\subsection{Adversarial Robustness (Classification)}
\label{sec:exp-adv}
We evaluate BITE's ability to improve model robustness for question answering and natural language understanding using SQuAD 2.0 \citep{rajpurkar-etal-2018-know} and MultiNLI \citep{mnli}, respectively. We use \mor\ \citep{morpheus20}, an adversarial attack targeting inflectional morphology, to test the overall system's robustness to non-standard inflections. They previously demonstrated \mor's ability to generate plausible and semantically equivalent adversarial examples resembling L2 English sentences. We attack each BERT$_{\text{base}}$ model separately and report F$_1$ scores on the answerable questions and the full SQuAD 2.0 dataset, following \citet{morpheus20}. In addition, for MNLI, we report scores for both the in-domain (MNLI) and out-of-domain dev.\ set (MNLI-MM).

\paragraph{BITE+WordPiece vs.\ only WordPiece.} First, we demonstrate the effectiveness of BITE at making the model robust to inflectional adversaries. After fine-tuning two separate BERT$_{\text{base}}$ models on SQuAD 2.0 and MultiNLI, we generate adversarial examples for them using \mor. From \Cref{tab:adv_robustness}, we observe that the BITE+WordPiece model not only achieves similar performance ($\pm$0.5) on clean data, but is significantly more robust to inflectional adversaries (10-point difference for SQuAD 2.0, 17-point difference for MultiNLI).

\paragraph{BITE vs.\ adversarial fine-tuning.}
Next, we compare the BITE to adversarial fine-tuning \citep{morpheus20}, an economical variation of adversarial training \citep{goodfellow2015} for making models robust to inflectional variation. 
In adversarial fine-tuning, an adversarial training set is generated by randomly sampling inflectional adversaries $k$ times from the adversarial distribution found by \mor\ and adding them to the original training set. Rather than retraining the model on this adversarial training set, the previously trained model is simply trained for one extra epoch. We follow the above methodology and adversarially fine-tune the WordPiece-only BERT$_{\text{base}}$ for one epoch with $k$ set to 4. To ensure a fair comparison, we also train the BITE+WordPiece BERT$_{\text{base}}$ on the original training set for an extra epoch.

From \Cref{tab:adv_robustness}, we observe that BITE is often more effective than adversarial fine-tuning at making the model more robust against inflectional adversaries and in some cases (SQuAD 2.0 All and MNLI-MM) even without needing the additional epoch of training. However, the adversarially fine-tuned model consistently achieves better performance on clean data. This is likely because even though adversarial fine-tuning requires only a single epoch of extra training, the process of generating the training set increases its size by a factor of $k$ and hence the number of updates. In contrast, BITE requires no extra training and is more economical.

Adversarial fine-tuning is also less effective at inducing model robustness when the adversarial example is from an out-of-domain distribution (8 point difference between MNLI and MNLI-MM). This makes it less useful for practical scenarios, where this is often the case. In contrast, BITE performs equally well on both in- and out-of-domain data, demonstrating its applicability to practical scenarios where the training and testing domains may not match. This is the result of preserving the base forms, which we investigate further in \Cref{subsec:robustness}.

\begin{table}[]
\small
    \centering
    \begin{tabular}{l c c c}
    \toprule
        \textbf{Condition} & \textbf{Encoding} & \textbf{BLEU} & \textbf{METEOR} \\
     \midrule
         \multirow{2}{*}{Clean} & BPE only & 29.13 & 47.80 \\
         & BITE + BPE & \textbf{29.61} & \textbf{48.31} \\
     \midrule
         \multirow{2}{*}{\mor} & BPE only & 14.71 & 39.54 \\
         & BITE + BPE & \textbf{17.77} & \textbf{41.58} \\
     \bottomrule
    \end{tabular}
    \vspace{-0.5em}
    \caption{Results on newstest2014 for Transformer-big trained on WMT'16 English-German (En-De).}
    \label{tab:nmt}
\end{table}

\subsection{Machine Translation}
\label{subsec:nmt}
Next, we evaluate BITE's impact on machine translation using the Transformer-big architecture \citep{ott2018-scaling} and WMT'14 English--German (En--De) task. We apply BITE+BPE to the English examples and compare it to the BPE-only baseline. More details about our experimental setup can be found in \Cref{app:nmt}.

To obtain the final models, we perform early-stopping based on the validation perplexity and average the last ten checkpoints. We observe that the BITE+BPE model converges 28\% faster (\Cref{fig:nmt_ppl}) than the baseline (20k vs.\ 28k updates) in addition to outperforming it by 0.48 BLEU on the standard data and 3.06 BLEU on the \mor\ adversarial examples (\Cref{tab:nmt}). This suggests that explicit encoding of morphological information helps models learn better and more robust representations faster.

\subsection{Dialectal Variation}
\label{sec:exp-dialect}

Apart from second languages, dialects are another common source of non-standard inflections. However, there is a dearth of task-specific datasets in English dialects like AAVE and CSE. Therefore, in this section's experiments, we use the model's pseudo perplexity (pPPL) \citep{wang2019pseudo} on monodialectal corpora as a proxy for its performance on downstream tasks in the corresponding dialect. The pPPL measures how certain the pretrained model is about its prediction and reflects its generalization ability on the dialectal datasets. To ensure fair comparisons across different subword segmentations, we normalize the pseudo log-likelihoods by the number of \emph{word tokens} fed into the WordPiece component of each tokenization pipeline \citep{Mie2016Can}. This avoids unfairly penalizing BITE for inevitably generating longer sequences. Finally, we scale the pseudo log-likelihoods by the masking probability (0.15) so that the final pPPLs are within a reasonable range.

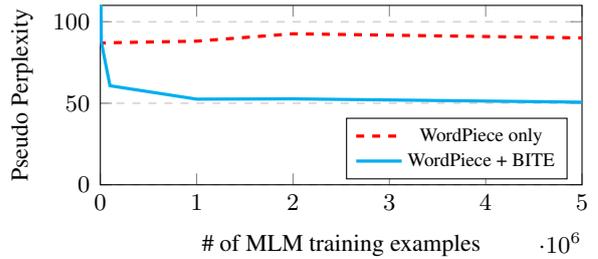
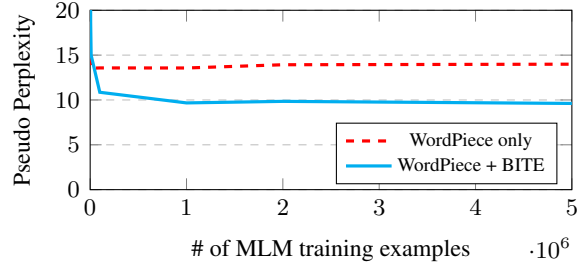
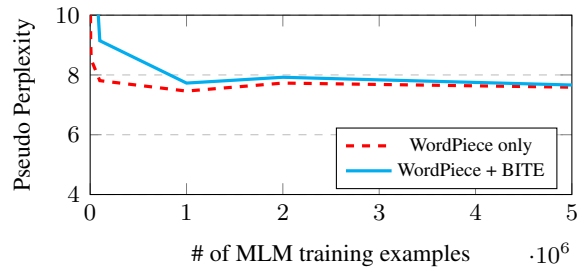
\begin{figure}[t]
    \centering
    \begin{subfigure}{0.495\textwidth}
        \centering
        \begin{tikzpicture}[font=\small]
        \begin{axis}[
            legend style={nodes={scale=0.8, transform shape}},
            width=\textwidth,
            height=0.5\textwidth,
            ymode=linear,
            xlabel={\# of MLM training examples},
            xtick scale label code/.code={\pgfmathparse{(#1)}$\cdot 10^{\pgfmathresult}$},
            ylabel = {Pseudo Perplexity},
            ylabel near ticks,
            xmin=0, xmax=5000000,
            ymin=0, ymax=110,
            legend pos=south east,
            ymajorgrids=true,
            grid style=dashed,
            every axis plot/.append style={no markers, very thick}
        ]
        
        \addplot[
            color=red,
            style=dashed
            ]
            coordinates { (1000,94.9672231532673)
(2000,96.8484941672247)
(5000,89.5123466015737)
(10000,86.8576383563367)
(100000,87.0866766617707)
(1000000,88.0719310250598)
(2000000,92.5830991838296)
(5000000,90.0602070721104)
            };
            \addlegendentry{WordPiece only}

        \addplot[
            color=cyan
            ]
            coordinates {
 (1000,331.894524027278)
(2000,177.405171357327)
(5000,113.771817364574)
(10000,87.5515549585091)
(100000,60.7233832348317)
(1000000,52.5090315942531)
(2000000,52.6761794488079)
(5000000,50.652866389742)
            };
            \addlegendentry{WordPiece + BITE}
        \end{axis}
        \end{tikzpicture}
        \caption{Colloquial Singapore English (forum threads)}
    \end{subfigure}{}
    \begin{subfigure}{0.495\textwidth}
    \centering
    \begin{tikzpicture}[font=\small]
    \begin{axis}[
        legend style={nodes={scale=0.8, transform shape}},
        width=\textwidth,
        height=0.5\textwidth,
        xlabel={\# of MLM training examples},
        xtick scale label code/.code={\pgfmathparse{(#1)}$\cdot 10^{\pgfmathresult}$},
        xmin=0, xmax=5000000,
        ymin=0, ymax=20,
        legend pos=south east,
        ymajorgrids=true,
        grid style=dashed,
        ylabel = {Pseudo Perplexity},
        ylabel near ticks,
        every axis plot/.append style={no markers, very thick}
    ]
    
    \addplot[
        color=red,
        style=dashed
        ]
        coordinates {
        (1000,14.7108814260994)
(2000,14.739182811553)
(5000,13.6378299123526)
(10000,13.536807123311)
(100000,13.5723038018344)
(1000000,13.5647491311553)
(2000000,13.9259174709704)
(5000000,13.9880773150121)
        };
        \addlegendentry{WordPiece only}

    \addplot[
        color=cyan,
        mark=triangle,
        ]
        coordinates {
        (1000,70.2475464676066)
(2000,31.5467103952565)
(5000,18.5894912063261)
(10000,14.9926346602318)
(100000,10.8563533723924)
(1000000,9.6713418700178)
(2000000,9.84313027639313)
(5000000,9.6117185595356)
        };
        \addlegendentry{WordPiece + BITE}
    \end{axis}
    \end{tikzpicture}
    \caption{African American Vernacular English (CORAAL)}
\end{subfigure}{}
\begin{subfigure}{0.495\textwidth}
        \centering
        \begin{tikzpicture}[font=\small]
        \begin{axis}[
            legend style={nodes={scale=0.8, transform shape}},
            width=\textwidth,
            height=0.5\textwidth,
            ymode=linear,
            xlabel={\# of MLM training examples},
            xtick scale label code/.code={\pgfmathparse{(#1)}$\cdot 10^{\pgfmathresult}$},
            ylabel = {Pseudo Perplexity},
            ylabel near ticks,
            xmin=0, xmax=5000000,
            ymin=4, ymax=10,
            legend pos=south east,
            ymajorgrids=true,
            grid style=dashed,
            every axis plot/.append style={no markers, very thick}
        ]
        
        \addplot[
            color=red,
            style=dashed
            ]
            coordinates {
 (1000,11.5917708827684)
(2000,10.5234560326447)
(5000,9.21875630931262)
(10000,8.517410496459)
(100000,7.80893758976303)
(1000000,7.46030403741812)
(2000000,7.72751529103245)
(5000000,7.5895560153203)
            };
            \addlegendentry{WordPiece only}

        \addplot[
            color=cyan
            ]
            coordinates {
 (1000,87.6647598828491)
(2000,39.0206802971368)
(5000,20.5141327531444)
(10000,13.9906323115202)
(100000,9.14466721932027)
(1000000,7.72764062737523)
(2000000,7.92162157120045)
(5000000,7.66318005414659)
            };
            \addlegendentry{WordPiece + BITE}
        \end{axis}
        \end{tikzpicture}
        \caption{Standard English (Wikipedia+BookCorpus)}
    \end{subfigure}{}
    \caption{Pseudo perplexity of BERT$_{\text{base}}$ on CSE, AAVE, Standard English corpora. BITE$_{abl}$ refers to the ablated version without grammatical information.
    }
    \vspace{-0.7em}
    \label{fig:dialect}
\end{figure}

\paragraph{Corpora.} For AAVE, we use the Corpus of Regional African American Language (CORAAL) \citep{kendall2018corpus}, which comprises transcriptions of interviews with African Americans born between 1891 and 2005. For our evaluation, only the interviewee's speech was used. In addition, we strip all in-line glosses and annotations from the transcriptions before dropping all lines with less than three words. After preprocessing, this corpus consists of slightly under 50k lines of text (1,144,803 word tokens, 17,324 word types).

To obtain a CSE corpus, we scrape the Infotech Clinics section of the Hardware Zone Forums\footnote{\href{https://forums.hardwarezone.com.sg}{forums.hardwarezone.com.sg}}, a forum frequented by Singaporeans and where CSE is commonly used. Similar preprocessing to the AAVE data yields a 2.2M line corpus (45,803,898 word tokens, 253,326 word types).

\paragraph{Setup.}
We take the same pretrained BERT$_{\text{base}}$ model and fine-tune two separate variants (with and without BITE) on English Wikipedia and BookCorpus \citep{ZhuBooks} using the masked language modeling (MLM) loss without the next sentence prediction (NSP) loss. We fine-tune for one epoch on increasingly large subsets of the dataset, since this has been shown to be more effective than doing the same number of gradient updates on a fixed subset \citep{2019t5}. Preprocessing steps are described in \Cref{app:cls}.

Next, we evaluate model pPPLs on the AAVE and CSE corpora, which we consider to be from dialectal distributions that differ from the training data which is considered to be Standard English. 
Since calculating the stochastic pPPL requires randomly masking a certain percentage of symbols for prediction, we also experiment with doing this for each sentence multiple times before averaging them. However, we find no significant difference between doing the calculation once or five times; the random effects likely canceled out due to the large sizes of our corpora. 

\paragraph{Results.}
From \Cref{fig:dialect}, we observe that the BITE+WordPiece model initially has a much higher pPPL on the dialectal corpora, before converging to 50--65\% of the standard model's pPPL as the model adapts to the presence of the new inflection symbols (e.g., \texttt{VBD}, \texttt{NNS}, etc.). Crucially, the models are not trained on dialectal corpora, which demonstrates the effectiveness of BITE at helping models better generalize to unseen dialects. For Standard English, WordPiece+BITE performs slightly worse than WordPiece, reflecting the results on QA and NLI in \Cref{tab:adv_robustness}. However, it is important to note that the WordPiece vocabulary used was not optimized for BITE; results from \Cref{subsec:nmt} indicate that training the data-driven tokenizer from scratch with BITE might improve performance.

\paragraph{CSE vs.\ AAVE.} Astute readers might notice that there is a large difference in pPPL between the two dialectal corpora, even for the same tokenizer combination. One possible explanation is that CSE differs significantly from Standard English in morphology and syntax due to its Austronesian and Sinitic influences \citep{tongue1974english}. In addition, loan words and discourse particles not found in Standard English like \emph{lah}, \emph{lor} and \emph{hor} are commonplace in CSE \citep{leimgruber2009singlish}. AAVE, however, generally shares the same syntax as Standard English due to its largely English origins \citep{poplack2000english} and is more similar linguistically. These differences are likely responsible for the significant increase in pPPL for CSE compared to AAVE.

Another possible explanation is that the BookCorpus may contain examples of AAVE since the BookCorpus' source, Smashwords, also publishes African American fiction. We believe the reason for the difference is a mixture of these two factors.

\subsection{Ablation Study}
\label{sec:disc}
To tease apart the effects of BITE's two components (lemmatization and inflection symbol) on task performance, we ablate the extra grammatical information from the encoding by replacing all inflection symbols with a dummy symbol (BITE$_\text{abl}$). As expected, BITE$_\text{abl}$ is significantly more robust to adversarial inflections (\Cref{tab:abl}) and the slight performance drop is likely due to the POS tagger being adversarially affected. However, different tasks likely require different levels of attention to inflections and BITE allows the network to learn this for each task. For example, NLI performance on clean data is only slightly affected by the absence of morphosyntactic information, while MT and QA performance is more significantly affected.

In a similar ablation for the pPPL experiments, we find that both the canonicalizing effect of the base form and knowledge of each word's grammatical role contribute to the lower pPPL on dialectal data (\Cref{tab:likelihood} in the Appendix). We discuss this in greater detail in \Cref{app:ppl} and also report the pseudo log-likelihoods and per-symbol pPPLs in the spirit of transparency and reproducibility.

\begin{table}[t]
\small
    \centering
    \begin{tabular}{>{\quad}l@{\hskip 1em}c c c c}
    \toprule
        & \multicolumn{2}{c}{\textbf{Clean}} & \multicolumn{2}{c}{\textbf{\mor}}\\
        \textbf{Dataset} & BITE$_\text{abl}$ & BITE & BITE$_\text{abl}$ & BITE \\
      \midrule
      \rowgroup{SQuAD 2 (F$_1$)}\\
        Ans. Qns. & 68.85 & \textbf{74.50} & 70.68 & \textbf{71.33}\\
        All Qns. & \textbf{72.90} & 72.71 & \textbf{69.29} & 69.23\\
      \midrule
      \rowgroup{MNLI (Acc.)} \\
        Matched & 82.28 & \textbf{83.01} & \textbf{80.17} & 76.11\\
        Mismatched & 83.18 & \textbf{83.50} & \textbf{81.21} & 76.64\\
      \midrule
        \rowgroup{WMT'14 (BLEU)} & 28.14 & \textbf{29.61} & \textbf{20.91} & 17.77\\
     \bottomrule
    \end{tabular}
    \caption{Effect of reinjecting grammatical information via inflection symbols. BITE$_\text{abl}$ refers to the ablation with the dummy symbol instead of inflection symbols.}
    \label{tab:abl}
\end{table}

\section{Model-Independent Analyses}
Finally, we analyze WordPiece, BPE, and unigram LM subword tokenizers that are trained with and without BITE. Implementation details can be found in \Cref{app:mia}. Through our experiments, we explore how BITE improves adversarial robustness and helps the data-driven tokenizer use its vocabulary more efficiently. We use 1M examples from Wikipedia+BookCorpus for training.

\subsection{Vocabulary Efficiency}
We may operationalize the question of whether BITE improves vocabulary efficiency in numerous ways. We discuss two vocabulary-level measures here and a sequence-level measure in \Cref{app:vocab_efficiency}.

\begin{figure}[ht]
    \centering
        \begin{tikzpicture}[font=\small]
        \begin{axis}[
        legend style={nodes={scale=0.85, transform shape}},
            legend pos=south east,
            width=0.5\textwidth,
            height=0.25\textwidth,
            ymode=linear,
            xlabel={Vocabulary Size (symbols) $\cdot 10^4$},
            xmin=0,
            xmax=20000,
           ylabel={Coverage (\%)},
        xtick scale label code/.code={},
                   ylabel near ticks,
            ymajorgrids=true,
            grid style=dashed,
            every axis plot/.append style={no markers, very thick}
        ]

        \addplot[
            color=red,
            style=dashed
            ]
            coordinates {
            ( 500 , 62.28657420200565 )
( 1000 , 68.62501542407786 )
( 3000 , 78.91308955509872 )
( 5000 , 83.47222276716406 )
( 10000 , 88.76243324807417 )
( 20000 , 92.72973228124775 )
( 30000 , 94.48071043402734 )
( 40000 , 95.50292543009856 )
( 50000 , 96.17652223537468 )
( 60000 , 96.65537529413847 )
( 70000 , 97.01533783915255 )
( 80000 , 97.29879655759404 )
( 90000 , 97.52809784546551 )
            };
        \addlegendentry{Baseline}

        \addplot[
            color=teal,
            ]
            coordinates {
            ( 500 , 69.63908495308468 )
( 1000 , 75.79135155909749 )
( 3000 , 84.90313562976073 )
( 5000 , 88.44052183751621 )
( 10000 , 92.22596885881266 )
( 20000 , 94.88828650288677 )
( 30000 , 96.04500089390267 )
( 40000 , 96.72153354585939 )
( 50000 , 97.17101405239322 )
( 60000 , 97.49511699695292 )
( 70000 , 97.74335511350131 )
( 80000 , 97.94022750583254 )
( 90000 , 98.10351488437018 )
            };
        \addlegendentry{BITE}
        \end{axis}
        \end{tikzpicture}
    \caption{Comparison of coverage between BITE and a trivial baseline (word counts). 
    }
    \label{fig:vocab_efficiency}

\end{figure}
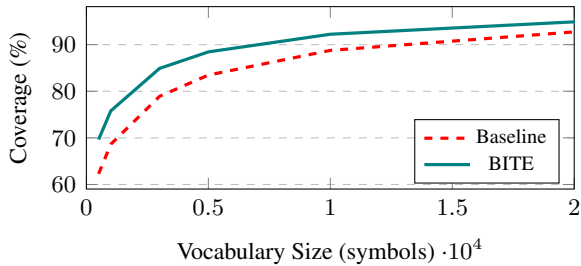

\paragraph{Vocabulary coverage.}
One measure of vocabulary efficiency is the coverage of a representative corpus by a vocabulary's symbols. We measure coverage by computing the total number of tokens (words and punctuation) in the corpus that are represented in the vocabulary divided by the total number of tokens in the corpus. We use the 1M subset of Wikipedia+BookCorpus as our representative corpus. Since BITE does not require a vocabulary size to be fixed before training, we set the $N$ most frequent types (base forms and inflections) to be our vocabulary. We use the $N$ most frequent types in the unencoded text as our baseline vocabulary.

From \Cref{fig:vocab_efficiency}, we observe that the BITE vocabulary achieves a higher coverage of the corpus than the baseline, hence demonstrating the efficacy of BITE at improving vocabulary efficiency. Additionally, we note that this advantage is most significant (5--7\%) when the vocabulary contains less than 10k symbols. This implies that inflected word forms comprise a large portion of frequently occurring types, which comports with intuition.

\paragraph{Symbol complexity.}
Another measure of vocabulary efficiency is the total number of symbols needed to encode a representative set of word types. We term this the \emph{symbol complexity}. Formally, given $N$, the total number of word types in the evaluation corpus; $S_i$, the sequence of symbols obtained from encoding the $i$th type; and $u_i$, the number of unknown symbols in $S_i$, we define:

\vspace{-0.65em}
\begin{equation}\label{eqn:avg_tok_comp}
   \text{SymbComp}(S_1, \ldots, S_N) = \sum^{N}_{i=1} |S_i|+u_i.
\end{equation}{}
\vspace{-0.5em}

While not strictly necessary when comparing vocabularies on the same corpus, normalizing \Cref{eqn:avg_tok_comp} by the number of word types in the corpus may be helpful for cross-corpus comparisons. For simplicity, we define 
the penalty of each extra unknown symbol to be double that of a symbol in the vocabulary.\footnote{$|S|$ contributes the extra count.} A general form of \Cref{eqn:avg_tok_comp} is in \Cref{app:mia}.

\begin{figure}[t!]
    \centering
    \begin{tikzpicture}[font=\small]
    \begin{axis}[
        legend style={nodes={scale=0.8, transform shape}},
        legend pos=north east,
        width=0.5\textwidth,
        height=0.35\textwidth,
        xlabel={Vocabulary Size (symbols) $\cdot 10^4$},
        xmin=0,
        xmax=40000,
        ymajorgrids=true,
        ylabel={Symbol Complexity $\cdot 10^5$},
        ylabel near ticks,
        xlabel near ticks,
        grid style=dashed,
        every axis plot/.append style={ very thick},
        xtick scale label code/.code={},
        ytick scale label code/.code={},
        ]
\addplot[
color=orange,
style=dashed,
opacity=0.8
]
coordinates {
(3000, 365305)
(4000, 334716)
(5000, 317799)
(10000, 279501)
(20000, 244711)
(30000, 227057)
(40000, 215351)
};
\addlegendentry{BPE}

\addplot[
color=orange,
opacity=0.8
]
coordinates {
(3000, 370925)
(4000, 336950)
(5000, 319777)
(10000, 276243)
(20000, 241808)
(30000, 224187)
(40000, 212651)
};
\addlegendentry{BPE + BITE}
        
\addplot[
color=violet,
style=dashed,
opacity=0.8 
]
coordinates {
(3000, 341896)
(4000, 316340)
(5000, 299822)
(10000, 261089)
(20000, 225759)
(30000, 206696)
(40000, 193968)
        };
\addlegendentry{WordPiece}
        
\addplot[
color=violet,
opacity=0.8 
]
coordinates {
(3000, 341570)
(4000, 315110)
(5000, 297396)
(10000, 254940)
(20000, 218396)
(30000, 200454)
(40000, 188555)
        };
\addlegendentry{WordPiece + BITE}

\addplot[
color=cyan,
style=dashed,
opacity=0.8 
]
coordinates {
(3000, 334221)
(4000, 321258)
(5000, 311874)
(10000, 280004)
(20000, 250073)
(30000, 232456)
(40000, 220378)
};
\addlegendentry{Unigram LM}

\addplot[color=cyan, opacity=0.8 ]
coordinates {
(3000, 333549)
(4000, 320638)
(5000, 309952)
(10000, 274913)
(20000, 244242)
(30000, 227525)
(40000, 216428)
};
\addlegendentry{Unigram LM + BITE}
        \end{axis}
        \end{tikzpicture}
    \caption{Symbol complexities of tokenizer vocabularies as computed in \Cref{eqn:avg_tok_comp}. Lower is better.}
    \label{fig:sem_capacity}
\end{figure}
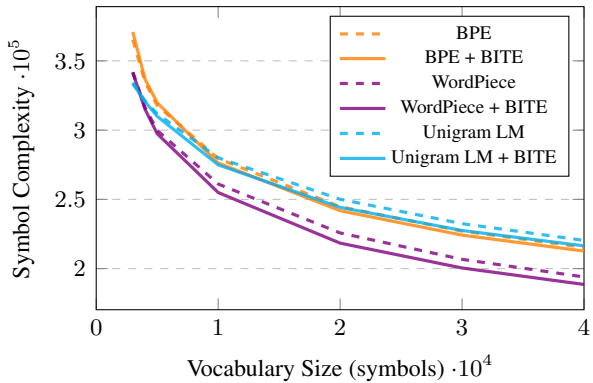

To measure the symbol complexities of our vocabularies, we use WordNet's single-word lemmas \citep{Miller95wordnet} as our ``corpus" ($N=83118$). From \Cref{fig:sem_capacity}, we see that training data-driven tokenizers with BITE produces vocabularies with lower symbol complexities.
Additionally, we observe that tokenizer combinations incorporating WordPiece or unigram LM generally outperform the BPE ones. We believe this to be the result of using a language model to inform vocabulary creation. It is logical that a symbol that maximizes a language model's likelihood on the training data is also semantically ``denser", hence prioritizing such symbols produces efficient vocabularies. We leave the in-depth investigation of this relationship to future work.

\subsection{Adversarial Robustness} \label{subsec:robustness}
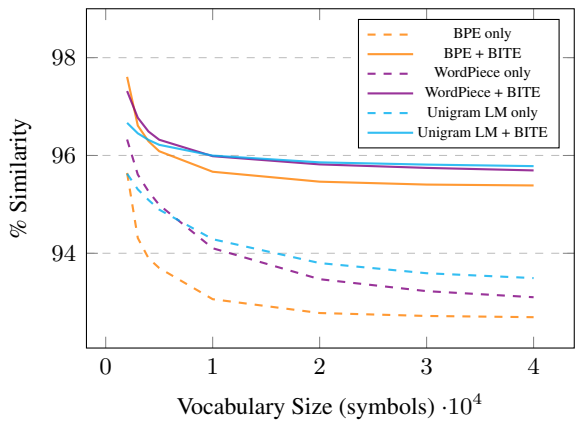
\begin{figure}[t!]
    \centering
\begin{tikzpicture}[font=\small]
        \begin{axis}[
            legend style={nodes={scale=0.63, transform shape}},
            legend pos=north east,
            width=0.5\textwidth,
            height=0.38\textwidth,
            xlabel={Vocabulary Size (symbols) $\cdot 10^4$},
            ymax=99,
        xtick scale label code/.code={},
     ymajorgrids=true,
            ylabel={\% Similarity},
            ylabel near ticks,
            xlabel near ticks,
            grid style=dashed,
            every axis plot/.append style={no markers,  thick}
        ]
\addplot[color=orange,
        style=dashed,
opacity=0.8]
        coordinates {
( 2000 , 95.6431830667632 )
( 3000 , 94.31195894877472 )
( 4000 , 93.89842838223984 )
( 5000 , 93.69864090924796 )
( 10000 , 93.06129027240205 )
( 20000 , 92.7782831817948 )
( 30000 , 92.71826538330302 )
( 40000 , 92.69311850764494 )
        };
\addlegendentry{BPE only}
\addplot[
        color=orange,
opacity=0.8
        ]
        coordinates {
( 2000 , 97.60749844037603 )
( 3000 , 96.60987578261752 )
( 4000 , 96.29012468894153 )
( 5000 , 96.08691301090126 )
( 10000 , 95.66689873765426 )
( 20000 , 95.46407918870744 )
( 30000 , 95.40467988551019 )
( 40000 , 95.38686766715372 )
        };
\addlegendentry{BPE + BITE}

\addplot[color=violet,
style=dashed,
opacity=0.8]
coordinates {
( 2000 , 96.3261406767852 )
( 3000 , 95.60916089611021 )
( 4000 , 95.26587833080976 )
( 5000 , 94.9890429698319 )
( 10000 , 94.10315902168364 )
( 20000 , 93.4700715645446 )
( 30000 , 93.22556209667543 )
( 40000 , 93.101241535394 )
        };
\addlegendentry{WordPiece only}
\addplot[color=violet,opacity=0.8 ]
        coordinates {
( 2000 , 97.31685540068241 )
( 3000 , 96.771635967056 )
( 4000 , 96.4894850617692 )
( 5000 , 96.32166041615292 )
( 10000 , 95.98513528419251 )
( 20000 , 95.8173390710784 )
( 30000 , 95.74580183629091 )
( 40000 , 95.69422030599213 )
        };
        \addlegendentry{WordPiece + BITE}

\addplot[
        color=cyan,
        style=dashed,
opacity=0.8
        ]
        coordinates {
( 2000 , 95.62651812936518 )
( 3000 , 95.30688938398892 )
( 4000 , 95.09001841564273 )
( 5000 , 94.89269184030103 )
( 10000 , 94.28793454138965 )
( 20000 , 93.79958940676934 )
( 30000 , 93.59181119631053 )
( 40000 , 93.49274325980102 )
        };
\addlegendentry{Unigram LM only}

\addplot[color=cyan,opacity=0.8 ]
        coordinates {
( 2000 , 96.66689636832443 )
( 3000 , 96.45253780067449 )
( 4000 , 96.31813659072986 )
( 5000 , 96.21746872817575 )
( 10000 , 95.99723475859778 )
( 20000 , 95.85896505920714 )
( 30000 , 95.8132710744401 )
( 40000 , 95.78193304684667 )
        };
\addlegendentry{Unigram LM + BITE}

        \end{axis}
        \end{tikzpicture}
    \caption{Mean percentage of symbols that are the same in the clean and adversarial encoded sequences.}
    \label{fig:clean_adv_sim}
\end{figure}
BITE's ability to make models more robust to inflectional variation can be directly attributed to its preservation of consistent, inflection-independent base forms. We demonstrate this by measuring the similarity between the encoded clean and adversarial sentences with the Ratcliff/Obershelp algorithm \citep{ratcliff1988pattern}. We use the MultiNLI in-domain development set and the \mor\ adversaries generated in \Cref{sec:exp-adv}.

We find that clean and adversarial sequences encoded by the BITE+$D$ tokenizers were more similar (1--2.5\%) than those encoded without BITE (\Cref{fig:clean_adv_sim}). 
The decrease in similarity with larger vocabularies is unsurprising; larger vocabularies result in shorter sequences, such that the same number of differing symbols will result in a larger relative change.

Hence, the improved robustness shown in \Cref{sec:exp-adv} can be directly attributed to the separation of each content word's base forms from its inflection and keeping it consistent as the inflection varies, hence mitigating any significant symbol-level changes. 

\subsection{Micro and Error Analysis}
\paragraph{Micro analysis.}
With a vocabulary of 20k symbols, BPE segments \textit{climbs} as $[$\textit{clim,bs}$]$, \textit{dreaming} as $[$\textit{dre,aming}$]$, and \textit{tumbled} as $[$\textit{t,umbled}$]$. WordPiece segments \textit{tumbled} as $[$\textit{tum,bled}$]$ and encodes \textit{dreaming} as a single symbol, but finds a morphologically accurate segmentation of \textit{climbs}: $[$\textit{climb,s}$]$. Unigram LM finds morphologically accurate segmentations for all three examples. When trained with BITE, all three tokenizers successfully find morphologically accurate segmentations of these examples and represent each corresponding base form as a single symbol.

\paragraph{Error analysis.}
Although the POS tagger is highly accurate\footnote{Accuracy of 97.2\% on the Wall Street Journal test set.}, it may occasionally tag an inflected form as a base form. An example from the MultiNLI data is the word \emph{turns} in \emph{``..., it could turns out even better"} being tagged as \texttt{NN} instead of \texttt{VBZ}. Consequently, this word would not be split into base form and inflection. Orthographic errors like misspellings also contribute to the tagger's inaccuracy. Some of these errors can be easily fixed by using a robust POS tagger \citep{moe}.

\section{Limitations}
Our BITE implementation relies on an external POS tagger to assign inflection tags to each word. This tagger requires language-specific training data, which can be a challenge for low resource languages. However, this could be an advantage since the overall system can be improved by training the tagger on dialect-specific datasets, or readily extended to other languages given a suitable tagger. Another drawback of BITE is that it increases the length of the encoded sequence which may lead to extremely long sequences if used on morphologically rich languages. However, this is not an issue for English Transformer models since the increase in length will \emph{always} be $<$2x, such that the increase in complexity is a constant factor.

\section{Conclusion}
\vspace{-0.3em}
The tokenization stage of the modern deep learning NLP pipeline has not received as much attention as the modeling stage, with researchers often defaulting to common subword tokenizers like BPE. We can do better.  By encoding raw text into operable symbols, we can improve the generalization and adversarial robustness of resulting systems.

Hence, we guide the data-driven tokenizer by incorporating linguistic information to learn a more efficient vocabulary and generate symbol sequences that increase the network's robustness to inflectional variation. This improves its generalization to L2 and World Englishes without requiring explicit training on such data. Since dialectal data is often scarce or even nonexistent, an NLP system's ability to generalize across dialects in a zero-shot manner is crucial for it to work well for diverse linguistic communities. A more general, BITE-like algorithm should enable further gains on morphologically rich languages.

Finally, given the effectiveness of the common task framework for spurring progress in NLP \citep{VarshneyKS2019}, we hope to do the same for tokenization. As a first step, we propose to evaluate an encoding scheme's efficacy by measuring its vocabulary coverage and symbol complexity (which may have interesting connections to information-theoretic limits \cite{ZivL1978}). We have already shown that Base-Inflection Encoding helps a data-driven tokenizer use its limited vocabulary more efficiently by reducing its symbol complexity when the combination is trained from scratch. 
\section*{Acknowledgments}
\vspace{-0.5em}
We are grateful to Michael Yoshitaka Erlewine from the NUS Dept.\ of English Language and Literature and our anonymous reviewers for their invaluable feedback. We also thank Xuan-Phi Nguyen for his help with reproducing the Transformer-big baseline. Samson is supported by Salesforce and Singapore's Economic Development Board under its Industrial Postgraduate Programme.

\bibliography{bibs/emnlp2020,bibs/nlp,bibs/adversarial,bibs/ling,bibs/misc,bibs/fairness}
\bibliographystyle{acl_natbib}
\clearpage
\appendix
\section{Examples of Inflectional Variation in English Dialects}
\label{app:dialect}

\paragraph{African American Vernacular English\\}
\citep{kendall2018corpus}
\begin{itemize}
    \item I dreamed about we \hl{was} over my uh, father mother house, and then we \hl{was} moving.
    \item I \hl{be} over with my friends.
    \item And this boy \hl{name} \texttt{RD-NAME-3}, he was \hl{tryna} be tricky, pretend like he \hl{don't} do nothing all the time.
\end{itemize}{}

\paragraph{Colloquial Singapore English (Singlish)\\} (Source: \href{https://forums.hardwarezone.com.sg}{forums.hardwarezone.com.sg}) 
\begin{itemize}
    \item Anyone face the problem after fresh \hl{installed} the Win 10 Pro, under NetWork File sharing after you enable this function (Auto discovery), the computer still failed to detect our Users connected to the same NetWork?
    \item I have \hl{try} it already, but no solutions appear.
    \item How \hl{do} time machine \hl{works}??
\end{itemize}{}

\section{Implementation/Experiment Details}
\label{app:impl_details}
\noindent All models are trained on 8 16GB Tesla V100s.
\begin{figure}[h]
    \centering
    \includegraphics[width=0.49\textwidth]{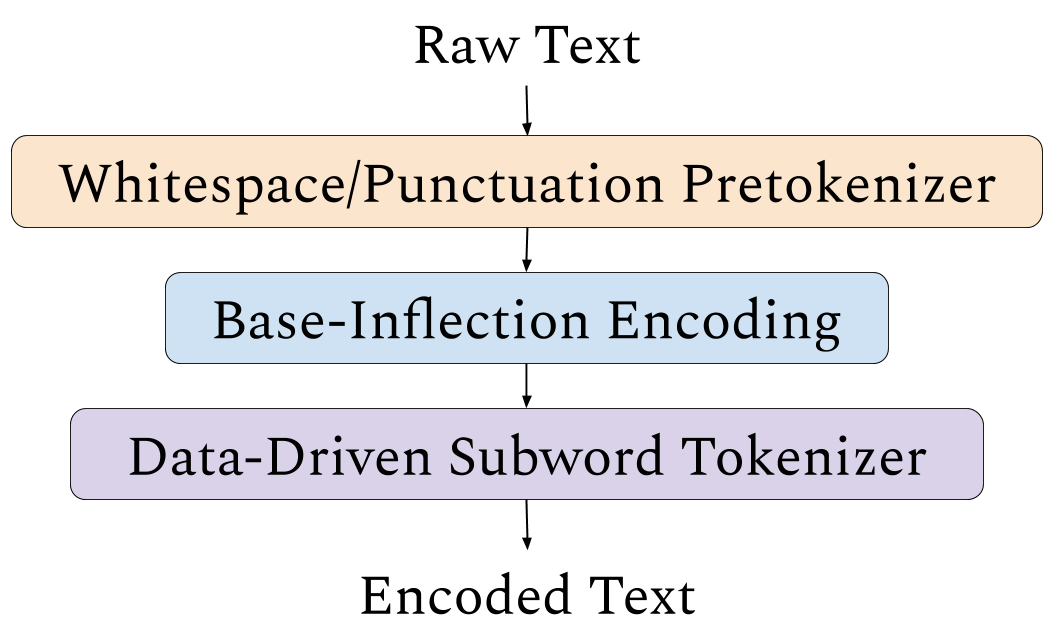}
    \caption{How BITE fits into the tokenization pipeline.}
    \label{fig:bite_pipe}
\end{figure}

\subsection{Classification Experiments}
\label{app:cls}
For our BERT experiments, we build BITE on top of the \texttt{BertTokenizer} class in \citet{Wolf2019HuggingFacesTS} and use their BERT implementation and fine-tuning scripts\footnote{\href{https://github.com/huggingface/transformers/tree/master/examples}{github.com/huggingface/transformers/.../examples}}. BERT$_\text{base}$ has 110M parameters. We do not perform a hyperparameter search and instead use the example hyperparameters for the respective scripts. 

\paragraph{Datasets and metrics.} MultiNLI \citep{mnli} is a natural language inference dataset of 392,702 training examples, 10k in-domain and 10k out-of-domain dev.\ examples, and 10k in-domain and 10k out-of-domain test examples spanning 10 domains. Each example comprises a premise, hypothesis, and a label indicating whether the premise entails, contradicts, or is irrelevant to the hypothesis. Models are evaluated using  $\text{Accuracy}=\frac{\text{\# correct predictions}}{\text{\# predictions}}$.

SQuAD 2.0 \citep{rajpurkar-etal-2018-know} is an extractive question answering dataset comprising more than 100k answerable questions and 50k unanswerable questions (130,319 training examples, 11,873 development examples, and 8,862 test examples). Each example is composed of a question, a passage, and an answer. Answerable questions are questions that can be answered by a span in the passage and unanswerable questions are questions that cannot be answered by a span in the passage. Models are evaluated using the $F_1$ score.

Wikipedia+BookCorpus is a combination of English Wikipedia and BookCorpus. We use \citet{lample2019cross}'s script to download and preprocess the Wikipedia dump before removing blank lines, overly short lines (less than three words or four characters), and lines with \texttt{doc} tags. We also remove blank and overly short lines from BookCorpus before concatenating and shuffling both datasets.

\subsection{Discussion for Perplexity Experiments}
\label{app:ppl}
\begin{table*}[h]
\small
    \centering
    \begin{tabular}{>{\quad}l c c c | c}
    \toprule
        & \textbf{WordPiece (WP)} & \textbf{WP + \textsc{Lemm}}  & \textbf{WP + BITE} & \textbf{WP + BITE$_{abl}$} \\
        \textbf{Dataset} & --- & (Lemmatize) & (+Infl. Symbols) & (+Dummy Symbol) \\
      \midrule
      \rowgroup{\textbf{Colloquial Singapore English}} & & & \\
        Total word tokens before WP & 45803898 & 45803898 & 51982873 & 51982873  \\
        Pseudo Negative Log-Likelihood & 30910290 & 30558864 & 31110740 & 30292923  \\
        pPPL (per word token before WP) & 92.58 & 85.43 & \textbf{52.67} & 48.66  \\
        pPPL (per symbol after WP) & 49.10 &	46.39  &	\textbf{32.02} & 30.20 \\
      \midrule
      \rowgroup{\textbf{African American Vernacular English}} & & & \\
        Total word tokens before WP & 1144803 & 1144803 & 1320730 & 1320730 \\
        Pseudo Negative Log-Likelihood & 452269 & 444021 & 453031 & 434621 \\
        pPPL (per word token before WP) & 13.92 & 13.27 & \textbf{9.84} & 8.96 \\
        pPPL (per symbol after WP) & 12.90 & 12.41 & \textbf{9.18} & 8.43 \\	
        \midrule
      \rowgroup{\textbf{Standard English}} & & & \\
        Total word tokens before WP & 252153 & 252153 & 290391 & 290391 \\
        Pseudo Negative Log-Likelihood & 77339 & 78074 & 90148 & 75467 \\
        pPPL (per word token before WP) & \textbf{7.72} & 7.87 & 7.92 & 5.65 \\
        pPPL (per symbol after WP) & 6.34 & 6.36 & \textbf{6.07} & 4.86\\
     \bottomrule
    \end{tabular}
    \vspace{-0.5em}
    \caption{Effect of lemmatization, inflection symbols, and dummy symbol on pseudoperplexity (pPPL). We also show the effect of normalizing by the word token vs. subword symbol count. Lower is better. \textbf{Bolded} values indicate lowest row-wise pPPLs, \emph{excluding} WP+BITE$_{abl}$ due to the confounding effect of the highly predictable dummy symbols.}
    \label{tab:likelihood}
\end{table*}
\paragraph{Effect of lemmatization and inflection symbols.}
We conduct two ablations to investigate the effects of lemmatization and inflection symbols on the models' pseudo perplexities: the first simply lemmatizes the input before encoding it with WordPiece (WordPiece+\textsc{Lemm}) and the second replaces every inflection symbol generated by BITE with a dummy symbol (WordPiece+BITE$_{abl}$). The latter is the same ablation used in \Cref{tab:abl} and from \Cref{tab:likelihood}, we see that this condition consistently achieved the lowest pPPL on all three corpora. However, we believe that the highly predictable dummy symbols likely account for the significant drops in pseudo perplexity.

To test this hypothesis, we perform another ablation, WordPiece+\textsc{Lemm}, where the the dummy symbols are removed entirely. If the dummy symbols were not truly responsible for the large drops in pPPL, we should observe similar results for both WordPiece+\textsc{Lemm} and WordPiece+BITE$_{abl}$. From \Cref{tab:likelihood} (pPPL per word token before WP), we see that the decrease in pPPL between WordPiece+\textsc{Lemm} and WordPiece is less drastic, thereby lending evidence for rejecting the null hypothesis.

\paragraph{Poorer performance on Standard English.} We observe that lemmatizing all content words and/or reinjecting the grammatical information appears to have the opposite effect on Standard English data compared to the dialectal data. Intuitively, such an encoding \emph{should} result in even more significant reductions in perplexity on Standard English since the POS tagger and lemmatizer were trained on Standard English data. A possible explanation for these results is that the WordPiece tokenizer and BERT model are overfitted on Standard English, since they were both (pre-)trained on Standard English data.

\paragraph{Normalizing log-likehoods.} In an earlier version of this paper, we computed pseudo perplexity by normalizing the pseudo log-likehoods with the number of masked subword symbols (the default). A reviewer pointed out that per subword symbol perplexities are not directly comparable across different subword segmentations/vocabularies, but per word perplexities are \citep{Mie2016Can,salazar-etal-2020-masked}.  However, using the same denominator would unfairly penalize models using BITE since it inevitably increases the symbol sequence length, which affects the predicted log-likelihoods. In addition, with the exception of the inflection/dummy symbols that replaced some unused tokens, the vocabularies of all the WordPiece tokenizers used in our pseudo perplexity experiments are exactly the same since we do not retrain them. Therefore, we attempt to balance these two factors by normalizing by the number of \emph{word tokens} fed into the WordPiece component of each tokenization pipeline in \Cref{fig:dialect}. We also report the per subword pPPL and raw pseudo negative log-likelihood in \Cref{tab:likelihood}.

\subsection{Machine Translation Experiments}
\label{app:nmt}
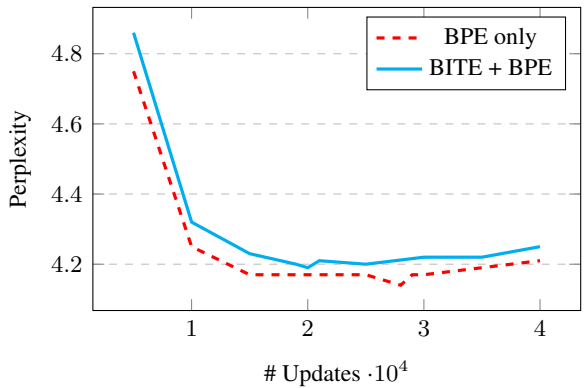
\begin{figure}[h]
    \centering
    \begin{tikzpicture}[font=\small]
    \begin{axis}[
        width=0.5\textwidth,
        height=0.35\textwidth,
        xlabel={\# Updates $\cdot 10^4$},
        xtick scale label code/.code={},
        legend pos=north east,
        ymajorgrids=true,
        grid style=dashed,
        ylabel = {Perplexity},
        ylabel near ticks,
        every axis plot/.append style={no markers, very thick}
    ]
    
    \addplot[
        color=red,
        style=dashed
        ]
        coordinates {
        (5000,4.75)
        (10000,4.25)
        (15000,4.17)
        (20000,4.17)
        (25000,4.17)
        (26000,4.16)
        (27000,4.15)
        (28000,4.14)
        (29000,4.17)
        (30000,4.17)
        (35000,4.19)
        (40000,4.21)
        };
        \addlegendentry{BPE only}

    \addplot[
        color=cyan,
        mark=triangle,
        ]
        coordinates {
        (5000,4.86)
        (10000,4.32)
        (15000,4.23)
        (19000,4.2)
        (20000,4.19)
        (21000,4.21)
        (25000,4.2)
        (30000,4.22)
        (35000,4.22)
        (40000,4.25)
        };
        \addlegendentry{BITE + BPE}
    \end{axis}
    \end{tikzpicture}
    \caption{Validation perplexity over the course of training for Transformer-big.}
    \label{fig:nmt_ppl}
\end{figure}
For our Transformer-big experiments, we use the \texttt{fairseq} \citep{ott2019fairseq} implementation and the hyperparameters from \citet{ott2018-scaling}:
\begin{itemize}[nosep]
    \item Parameters: 210,000,000
    \item BPE operations: 32,000
    \item Learning rate: 0.001
    \item Per-GPU batch size: 3,584 tokens
    \item Warmup period: 4,000 updates
    \item Dropout: 0.3
    \item Gradient Accumulation: 16
\end{itemize}

Both models took 24.6 hours to complete 45k updates. We use a \texttt{fairseq} script\footnote{\href{https://github.com/pytorch/fairseq/blob/master/scripts/average_checkpoints.py}{github.com/pytorch/fairseq/.../average\_checkpoints.py}} to average the selected checkpoint with the previous nine.

\paragraph{Dataset and metrics.}
We use the WMT'16 data\footnote{\href{https://statmt.org/wmt16/translation-task.html}{statmt.org/wmt16/translation-task.html}} for training and newstest2013 for development and newstest2014 for testing. Although it is common practice to use the already encoded WMT'16 data released by Google, BITE requires raw or whitespace-tokenized text as input. Hence, we use the raw WMT data and the \texttt{fairseq} \href{https://github.com/pytorch/fairseq/blob/master/examples/translation/prepare-wmt14en2de.sh}{preprocessing script} to preprocess our data. After preprocessing, we obtain a dataset of 4.3M training examples, 2,996 dev. examples, and 3,003 test examples. Models are evaluated using BLEU\footnote{Calculated by \texttt{fairseq}.} \citep{bleu_score_papineni2002bleu} and METEOR\footnote{\href{http://www.cs.cmu.edu/~alavie/METEOR/}{cs.cmu.edu/~alavie/METEOR/}} \citep{meteor-wmt}, standard MT evaluation metrics.

\subsection{Model-Independent Analyses}
\label{app:mia}
 We use the \texttt{tokenizers} implementation of WordPiece and BPE and the SentencePiece \citep{kudo2018sentencepiece} implementation of unigram LM. For ease of comparison across the three encoding schemes, we pretokenize the raw text with \texttt{tokenizers} BertPreTokenizer before encoding them. For practical applications, users may use \texttt{sentencepiece}'s method of handling whitespace characters instead of the BertPreTokenizer.
 
\paragraph{General form of \Cref{eqn:avg_tok_comp}.}
\begin{equation}
   \text{SymbComp}(S_1, \ldots, S_N) = \sum^{N}_{i=1} (|S_i|-u_i)+\lambda u_i
\end{equation}{}
\noindent where $N$ is the total number of word types in the evaluation corpus, $S_i$ is the sequence of symbols obtained from encoding the $i$th base form, $u_i$ is the number of unknown symbols in $S_i$, and $\lambda$ is the weight of the unknown symbol penalty.
 
\paragraph{Ratcliff/Obershelp algorithm.} We use Python's \texttt{difflib} implementation.

\section{More Measures of Vocabulary Efficiency}
\label{app:vocab_efficiency}
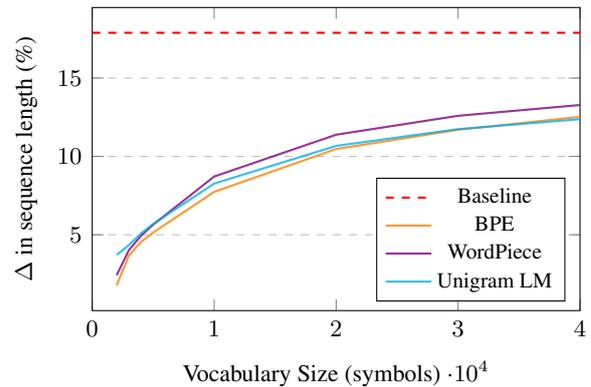
\begin{figure}[H]
    \centering
        \begin{tikzpicture}[font=\small]
        \begin{axis}[
        legend style={nodes={scale=0.9, transform shape}},
            width=0.5\textwidth,
            height=0.35\textwidth,
            ymode=linear,
            xmin=0,
            xmax=40000,
            xlabel={Vocabulary Size (symbols) $\cdot 10^4$},
           ylabel={$\Delta$ in sequence length (\%)},
        xtick scale label code/.code={},
                   ylabel near ticks,
            legend pos=south east,
            ymajorgrids=true,
            grid style=dashed,
            every axis plot/.append style={no markers, thick}
        ]
\addplot[
            color=red,
            style={dashed}
            ]
            coordinates {
            (1, 17.89)
            (40000, 17.89)
 };
\addlegendentry{Baseline}
 
\addplot[
color=orange, opacity=0.8            ]
coordinates {
(2000, 1.7714940591755082)
 (3000, 3.6703245793111794)
 (4000, 4.541372050286605)
 (5000, 5.144190855119906)
 (10000, 7.739708389027185)
 (20000, 10.469985722901628)
 (30000, 11.69819896596784)
 (40000, 12.528304081239526)
            };
\addlegendentry{BPE}

\addplot[
color=violet,
opacity=0.8
]
coordinates {
(2000, 2.4185648043211776)
 (3000, 4.012952943105438)
 (4000, 4.939986086047924)
 (5000, 5.64887099200032)
 (10000, 8.715955592974773)
 (20000, 11.385053653823618)
 (30000, 12.591780299418325)
 (40000, 13.280174327707735)
 };
 \addlegendentry{WordPiece}
            
\addplot[
color=cyan, opacity=0.8           ]
coordinates {
(2000, 3.708591598840384)
 (3000, 4.342970918687156)
 (4000, 5.088427880790639)
 (5000, 5.693641618497106)
 (10000, 8.26779131722552)
 (20000, 10.67125526017256)
 (30000, 11.739701953164067)
 (40000, 12.377618572594548)
            };
 \addlegendentry{Unigram LM}
            
        \end{axis}
        \end{tikzpicture}
        \vspace{-0.5em}
    \caption{Relative increase in mean encoded sequence lengths (\%) between BITE-less and BITE-equipped tokenizers after training the data-driven subword tokenizers with varying vocabulary sizes; lower is better. Baseline (dotted red line) denotes the percentage of inflected forms in an average sequence; this is equivalent to the increase in sequence length if BITE had no effect on the data-driven tokenizers' encoding efficiency.}
            \label{fig:mean_seq_len}
\end{figure}
\paragraph{Sequence lengths.} A possible concern with BITE is that it may significantly increase the length of the encoded sequence, and hence the computational cost for sequence modeling, since it splits all inflected content words (nouns, verbs, and adjectives) into two symbols. We calculate the percentage of inflected words to be 17.89\%.\footnote{Note that only content words are subject to inflection.} Therefore, if BITE did not enhance WordPiece's and BPE's encoding efficiency, we should expect a 17.89\% increase (i.e., upper bound) in their mean encoded sequence length. However, from \Cref{fig:mean_seq_len}, we see this is not the case as the relative increase (with and without BITE) in mean sequence length generally stays below 13\%, 5\% less than the baseline. This demonstrates that BITE helps the data-driven tokenizer make better use of its limited vocabulary. 

In addition, we see that the gains are inversely proportional to the vocabulary size. This is likely due to the following reasons. For a given sentence, the corresponding encoded sequence's length usually decreases as the data-driven tokenizer's vocabulary size increases as it allows merging of more smaller subwords into longer subwords. On the other hand, BITE is vocabulary-independent, which means that the encoded sequence length is always the same for a given sentence. Hence, the same absolute difference contributes to a larger relative increase as the vocabulary size increases. Additionally, more inflected forms are memorized as the vocabulary size increases, resulting in an average absolute increase of 0.4 symbols per sequence for every additional 10k vocabulary symbols. Together, these two factors explain the above phenomenon.

\end{document}